\documentclass[sigconf]{acmart} 

\AtBeginDocument{%
  \providecommand\BibTeX{{%
    \normalfont B\kern-0.5em{\scshape i\kern-0.25em b}\kern-0.8em\TeX}}}

\setcopyright{acmcopyright}
\copyrightyear{2021}
\acmYear{2021}
\acmDOI{10.1145/1122445.1122456}

\acmConference[]{}{}{}

\usepackage{amsmath}
\usepackage{amsthm}

\usepackage{amssymb}
\usepackage{booktabs}
\usepackage{graphicx}
\usepackage{mathtools}
\usepackage{algorithm}
\usepackage{algorithmic}

\newcommand{\eg}{\textit{e.g.}}
\newcommand{\ie}{\textit{i.e.}}

\usepackage{amsmath,amsfonts,bm}

\def\1{\bm{1}}

\def\vr{{\bm{r}}}
\def\vs{{\bm{s}}}

\def\vu{{\bm{u}}}

\def\vx{{\bm{x}}}
\def\vy{{\bm{y}}}
\def\vz{{\bm{z}}}

\DeclareMathAlphabet{\mathsfit}{\encodingdefault}{\sfdefault}{m}{sl}
\SetMathAlphabet{\mathsfit}{bold}{\encodingdefault}{\sfdefault}{bx}{n}

\newcommand{\E}{\mathbb{E}}

\newcommand{\R}{\mathbb{R}}

\newcommand{\KL}{D_{\mathrm{KL}}}

\renewcommand\footnotetextcopyrightpermission[1]{}
\usepackage{fancyhdr}
\usepackage{xcolor}

\acmSubmissionID{1975}

\begin{document}

\title{OR-Net: Pointwise Relational Inference for Data Completion under Partial Observation}



\author{Qianyu Feng}
\affiliation{
  \institution{University of Technology Sydney}
  \country{}}
  
\author{Linchao Zhu}
\affiliation{
  \institution{University of Technology Sydney}
  \country{}}
  
\author{Bang Zhang}
\affiliation{
  \institution{Alibaba DAMO Academy}
  \country{}}
  
\author{Pan Pan}
\affiliation{
  \institution{Alibaba DAMO Academy}
  \country{}}
  
\author{Yi Yang}
\affiliation{
  \institution{University of Technology Sydney}
  \country{}}



\renewcommand{\shortauthors}{}

\begin{abstract}
Contemporary data-driven methods are typically fed with full supervision on large-scale datasets which limits their applicability.
However, in the actual systems with limitations such as measurement error and data acquisition problems, people usually obtain incomplete data. Although data completion has attracted wide attention, the underlying data pattern and relativity are still under-developed. 
Currently, the family of latent variable models allows learning deep latent variables over observed variables by fitting the marginal distribution. 
As far as we know, current methods fail to perceive the data relativity under partial observation. Aiming at modeling incomplete data, this work uses relational inference to fill in the incomplete data. Specifically, we expect to approximate the real joint distribution over the partial observation and latent variables, thus infer the unseen targets respectively.
To this end, we propose Omni-Relational Network (OR-Net) to model the pointwise relativity in two aspects:
(i) On one hand, the inner relationship is built among the context points in the partial observation; 
(ii) On the other hand, the unseen targets are inferred by learning the cross-relationship with the observed data points.
It is further discovered that the proposed method can be generalized to different scenarios regardless of whether the physical structure can be observed or not. 
It is demonstrated that the proposed OR-Net can be well generalized for data completion tasks of various modalities, including function regression, image completion on MNIST and CelebA datasets, and also sequential motion generation conditioned on the observed poses.
\end{abstract}




\maketitle

\begin{figure}[t]
  \centering
  \vspace{6mm}
  \includegraphics[width=\columnwidth]{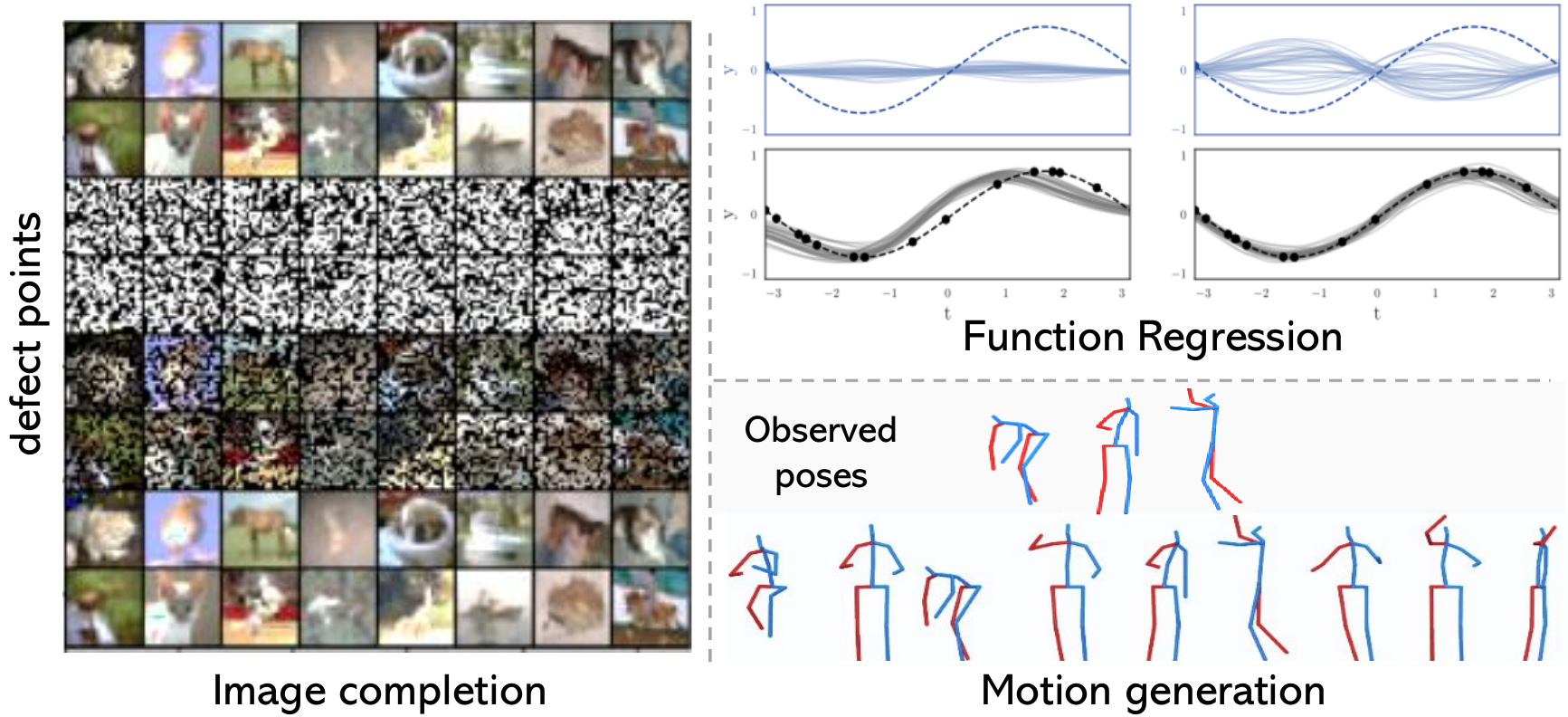}
  \caption{
    Different application scenarios for data completion with conditional generation under partial observation. Take image completion for example, it is widely applied when images are damaged with defect points. Besides, it is applicable for modeling the function in systems where only few variables can be measured. We further investigate the completion of motion sequences with several observed poses as the context.
  }
  \label{fig:teaser}
\end{figure}

\section{Introduction}

Deep learning has evolved the data-driven methods that rely on rich supervision in large-scale datasets~\cite{krizhevsky2012imagenet,lin2014coco,kriz2009cifar}. However, these methods suffer from over-fitting the training data and thus being vulnerable to attacks, which impedes the generalization ability. Stochastic processes~\cite{insua2012bayesian,lei2016new,ye2015stochastic,10.2307/1911358} provide alternative options to cover the distribution of data samples. 
With the estimated distribution conditioned on the context samples, the unseen targets can be inferred from the partial observation. A collection of models are devised to tackle these issues via learning a distribution over predictors, which allows incorporating data from observation and related tasks. 
However, traditional approaches are computationally expensive and intractable to be applied to large datasets. 
The family of neural processes~\cite{garnelo2018conditional,garnelo2018neural,kim2019attentive} are proposed to leverage the efficiency and scalability of neural network to parameterize the data distribution with a set of input-output pairs as exemplars. 
Nonetheless, they merely focus on learning the appearance features while overlooking the relativity among data samples, thus tend to fail to generalize to different modalities under various scenarios.

Differently, we aim to take a step further to learn the joint distribution over both observed and latent variables. It is challenging to learn the underlying joint distribution which assists model to approximate prior and posterior distributions over latent variables. Learning these distributions can be beneficial for various purposes, for example, inferring the latent distribution where data is originated. 
The critical limitation in the existing architectures~\cite{ye2015stochastic,garnelo2018neural,kim2019attentive} is that they fail to build induction across a broader context due to the unobserved heterogeneity. It is widely adopted that stacks of feature‐extracting layers compose the model, with the first layer processing raw inputs and the subsequent layers gradually learning more complex representation given a set of input/output pairs. \cite{kim2019attentive} tried to apply the attention mechanism into the neural processes to capture the linear mapping of data points. However, the linear attentive layers fail to satisfy diverse cases thus lead to blurry results for data completion.
Therefore, we consider modeling the relationship in data points with functions incorporating prior knowledge, \ie, position information, geometric structure, extracting from the neighboring context. 
Unlike convolutional neural network (CNN)~\cite{lecun1989backpropagation,krizhevsky2012imagenet} and recurrent neural network (RNN)~\cite{hochreiter1997long,elman1990finding} exploiting simple redundancies and invariances in data, GNN functions by directly encoding the structural information presented in data.
Long-range dependencies in the nonstructural data can be better captured via graph structures, allowing the information to propagate for multiple-hops hierarchically. 
An implication of this is that, among data points sampled from the distribution, they merely use the distance information to re-weight the importance of messages passed from the neighbors. This frequently leads to introduce blurring artifacts in incomplete or missing contexts.

Inspired by these works, our main idea is to leverage the structural relationship in data points for learning the target samples \textit{w.r.t.} other observed samples in the context set. To learn such embedding, we first connect each data point to its context points in the neighborhood, then learn the relative information for each pair. Taking in both the appearance embedding and geometric embedding, an attentive aggregation scheme further weighs the neighbors to perform the message passing. Such aggregations can be naturally chained and combined into multiple layers to enhance the expressiveness of representation. In the reconstruction of unseen target samples, the geometric embedding is inductive across different node orderings. When other attributes are available, the node embedding is further enriched by aggregating messages from its neighborhood. 

Therefore, the proposed OR-Net is equipped with the following desirable traits:
(i) \textbf{Structure-aware}: We devise a new realization of neural processes to incorporate the geometric information which is more expressive by capturing the relativity in the data points; 
(ii) \textbf{Reliability}: We propose to learn a compact representation via reducing the redundant information for learning an optimal estimation of data distribution; 
(iii) \textbf{Better Generalization}: OR-Net is directly applicable to data completion under different scenarios, \eg, 1D function, imperfect images and human motion sequences, where it can better estimate the unseen samples. 
Furthermore, we show that OR-Net is expressive and achieves competitive performances on multifold tasks, including function regression, image completion and motion generation.

\section{Related Work}

\textbf{Implicit Learning.}
Albeit the remarkable successes reached by deep learning over these years, the efficiency of learning~\cite{lake2015human,santoro2016one} on large datasets still remains a long-standing problem. One possible solution is to model function $f$ on the data distribution using only an arbitrary number of samples by exploiting the domain-wide statistics. Gaussian Processes~\cite{williams1996gaussian} capture the prior knowledge in the distributional assumptions with 
Bayesian inference. Conditioned on the observed samples, GP learns a parametric function $g$ approximating $f$ in the functional space where $g$ is initialized as a random function distributed according to the predictive posterior distribution. However, GP tends to fail to face the big data or large dimensionality~\cite{csato2002sparse,snelson2006sparse} due to the high intensive computation depending on the choice of kernels. 
Another prototype of the model is proposed by~\cite{garnelo2018neural}, named as Neural Processes. 
NP can be trained in an end-to-end style which combines neural networks with learning the approximating functions in a similar way as GP. 
The difference lies in that NP learns parameterized functions in a supervised way via gradient descent optimization which computes more efficiently by learning the uncertainties directly from the data. 
The regression on a toy 1D domain is discussed in~\cite{bachman2018vfunc} with a similar setting but optimizing an approximation to the entropy of the modeled function.
However, the aforementioned models are unconscious about the relationship among data samples. Without learning the relation in the data structure,  models tend to embed similar samples at different positions into the same embedding.
However, learning is still done in a GP framework by maximizing the marginal likelihood. The linear attention layers limit the expressiveness of the representation with the overlook in the inherited relationship in the set of data samples.

\noindent\textbf{Causal Inference.}
Learning the distribution conditioned on a randomly sampled subset, the NP-related task can also be regarded as few-shot learning. Most techniques cannot rapidly generalize from a few amount of examples relying on learning from large-scale data. However, the cost of computation will become a side-effect.
Meta-learning with neural networks is one example of this approach. Given input-output pairs drawn from a new function at test time, one can reason about this function by looking at the predictive distribution conditioning on these input-output pairs. There exist substantial works in few-shot classification~\cite{kipf2016semi,vinyals2016matching,snell2017prototypical,santoro2016one} dealing with locating relevant observed image/prototype given a query image. 
\cite{edwards2016towards,hewitt2018variational} have a similar permutation invariant encoder that predicts summaries of a data set using local latent features on top of global features. \cite{sung2018learning} learns to explore the relationship between samples a deep distance metric to compare a small number of images. \cite{garcia2017few}  studies the problem of few-shot learning with the prism of inference on a partially observed graphical model. 
Many approaches either assume full knowledge of the intervention, make strong assumptions about the model class, or have scalability limitations.

\noindent\textbf{Deep Variable Models.}
To some extent, Neural Processes tackle the problem from a representation learning~\cite{bengio2013representation}
perspective. It is ubiquitous that a stack of feature‐extracting layers compose the model, with the first layer processing raw inputs and the subsequent layers gradually learning more complex representation given a set of input/output pairs.
Unlike convolutional neural network (CNN)~\cite{lecun1989backpropagation,krizhevsky2012imagenet}and recurrent neural network (RNN)~\cite{hochreiter1997long,elman1990finding} exploiting simple redundancies and invariances in grid‐structured data, GNN models are more data‐efficient by directly encoding the structural inductive biases present in data. 
Existing GNN models belong to a family of graph architectures that adopt different aggregation schemes for learning the representation of the nodes.
Current models focus on learning representations that capture local network structure around a given node. Without relying on the node feature information, above models will always embed nodes at symmetric positions into same embedding vectors, which means that such nodes are indistinguishable from the GNN’s point of view.
Returning to our motivation for implanting position embedding in NPs, a clear parallel exists between position embedding learning and self-attention. Both of them targeting at digging into the relationship among data samples for generating valuable representations. Current models focus too much on learning node embeddings which tend to become assimilated.
So as to relieve this kind of issue, \cite{kipf2016semi,hamilton2017inductive} assign an unique node identifier with pre-trained transductive node features for training. 
However, such models are limited which cannot be generalized to unseen targets where the ordering or structure is unbeknownst. In contrast, PGNN~\cite{you2019position} propose to add up the positional information to the node embeddings in the graph.
Another option to assimilate the positional information is adopting a graph kernel that relies on the node positional information. Graph kernels implicitly or explicitly map graphs to a Hilbert space. Weisfeiler-Lehman and Subgraph kernels
have been incorporated into deep graph kernels~\cite{yanardag2015deep} to capture structural properties of neighborhoods. \cite{kashima2003marginalized,gartner2003graph} also propose to exploit graph kernels based on random walks, which count the number of walks two graphs have in common.

\section{Omni-Relational Network}\label{ss_OR-Net}

Aiming at measuring the underlying distribution, variational models are put forward to reconstruct the data space conditionally. However, without exploiting the relation among data points, it is intractable to learn reliable and generalized representations invariant to the given context points. 
To address the above issue, we propose Omni-Relational Network (OR-Net) to model the intrinsic data relationship. First, our network is built upon variational auto-encoders to estimate the joint distribution over the latent and observed variables. For better learning the data relativity, we not only build connections inner the context observed points, but also cross the given context and unseen targets. 
With the rich features and relativity learned, we further add constraint in the learning objectives with information bottleneck to guarantee the compactness of representations.

\begin{figure}[t]
  \centering
  \includegraphics[width=0.95\columnwidth]{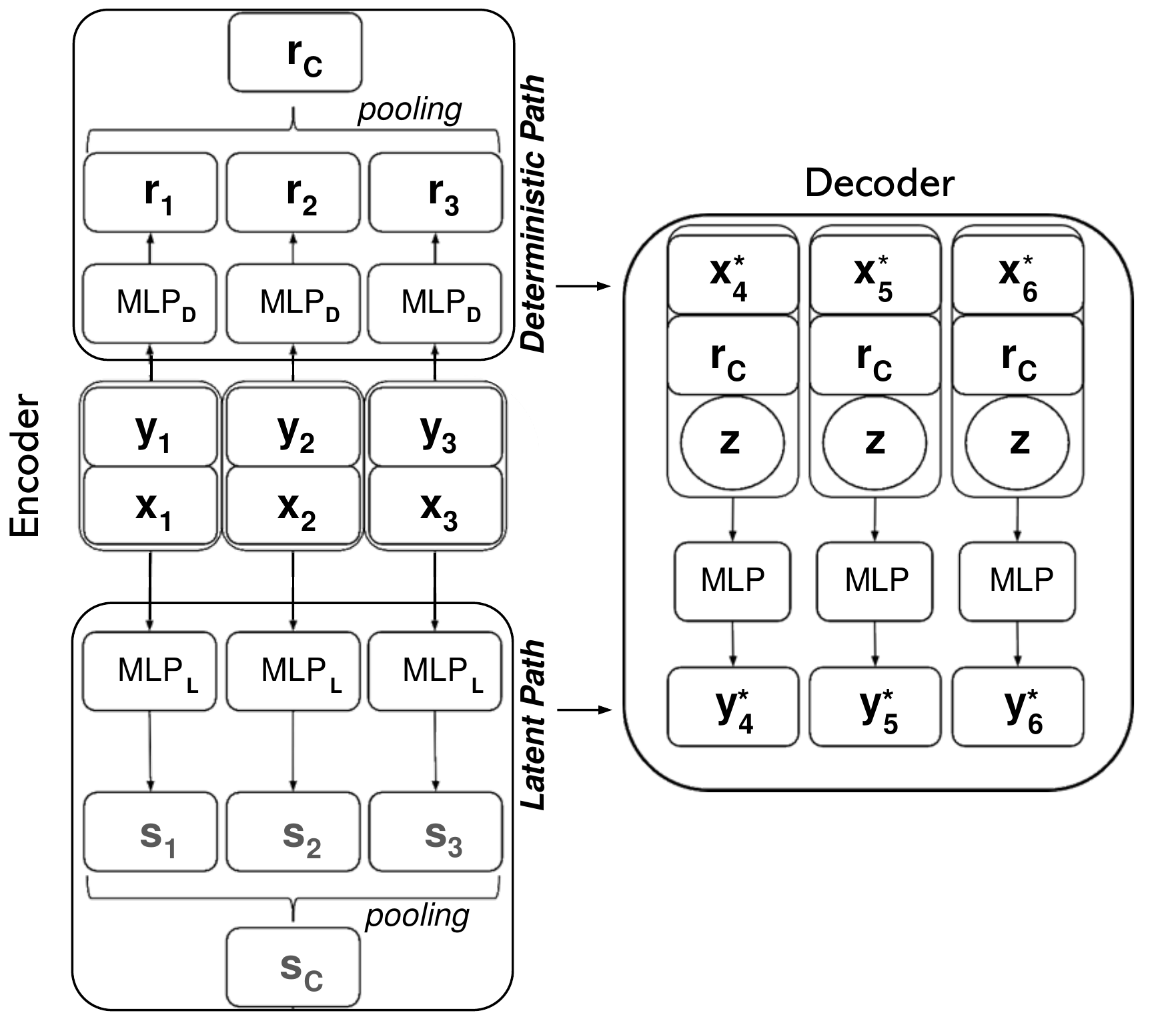}
  \caption{
     The network structure of Neural Processes is built on the encoder-decoder structure, where exist a deterministic path and latent path in the encoder for modeling the latent variables. $\vz$ is re-parameterized from $s_c$. 
  }
  \label{fig:framework1}
\end{figure}

\begin{figure*}[t]
  \centering
  \includegraphics[width=1.9\columnwidth]{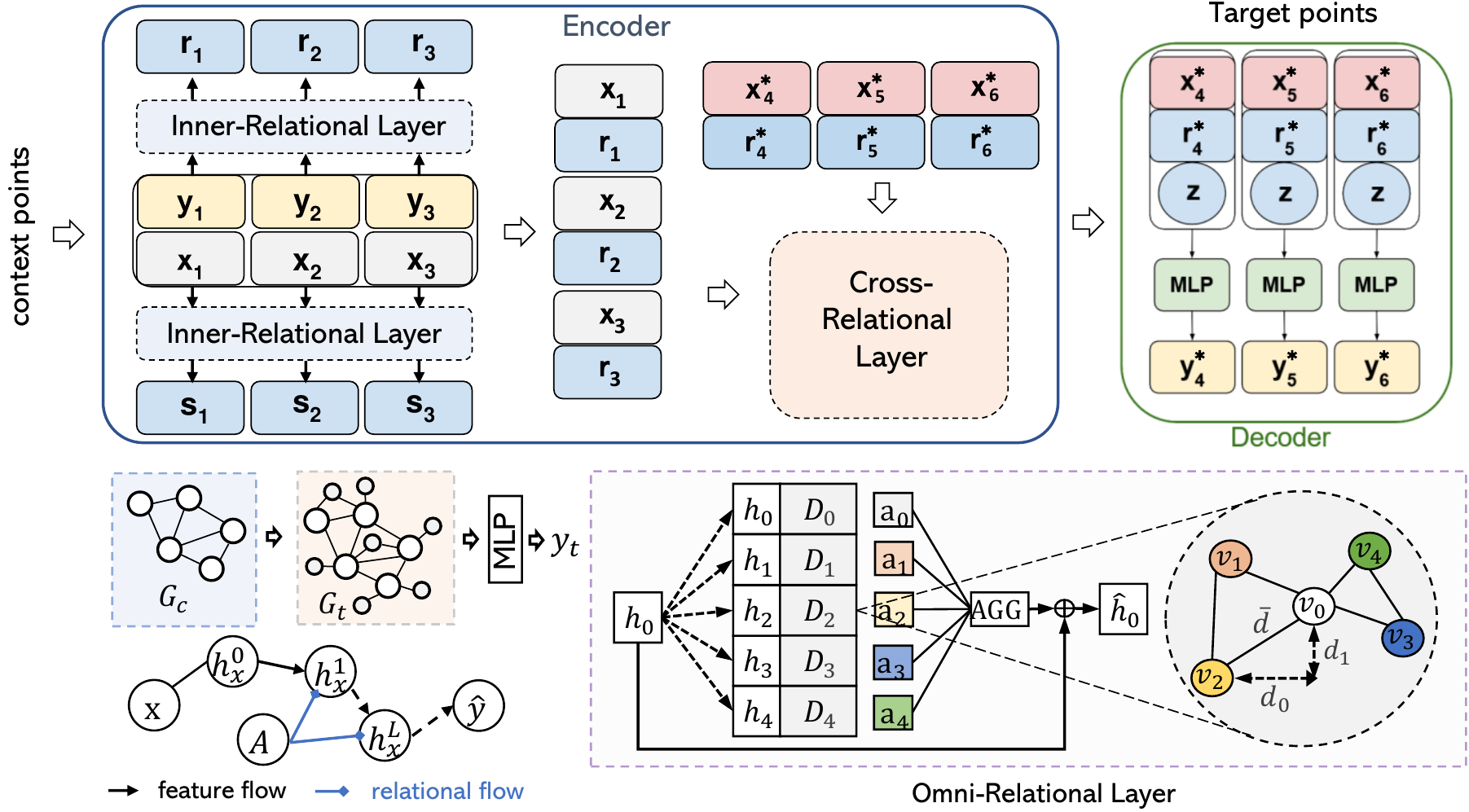}
  \caption{
     Illustration of the message passing in OR-Net. Given a point set, each sample is a node connected to other nodes within its neighborhood. Then, the embedding of the node feature together with the learned geometric embedding are messages to be attentively aggregated into the next layer. The right part shows the details about the geometric attentive message aggregation and updating. 
  }
  \label{fig:framework2}
\end{figure*}

\subsection{Preliminaries}

Stochastic processes aim to empirically fit the data distribution over the observed inputs to generate the values for unobserved samples.
Given the context samples $\{\vx_j,\vy_j\}_{j\in C}$, the goal is to learn a function to map an input $\vx_i \in \mathbb{R}^{d}$ to an output $\vy_i \in \mathbb{R}^{d}$ in the target space, $i\in T$. 
The data distribution can be modeled with parameterized functions learned on a set of observed samples which is invariant to the permutations of the elements. Here, we denote $\tau$ as a permutation for a set of $n$ samples, then: 
\begin{equation}
\begin{aligned}
    p(\vy_{1:n}|\vx_{1:n}) &\coloneqq p(\vy_1,...,\vy_n|\vx_1,...,\vx_n)\\
    &= p(\vy_{\tau(1)},...,\vy_{\tau(n)}|\vx_{\tau(1)},...,\vx_{\tau(n)})\\
    &\coloneqq p(\tau(\vy_{1:n})|\tau(\vx_{1:n})).
\end{aligned}
\end{equation}
Additionally, conditioned on the context set $C$, representations can be learned with a finite dimensionality which should also be consistent across context sets with different realizations of the data generation.
Modeled as regression functions, the aim is to estimate and optimise on a set of target samples $\{\vx_i,\vy_i\}_{i\in T}$:
\begin{equation}
    p(\vy_i|\vx_i,\vx_C,\vy_C) = p(\vy_i|\vx_i,\vr_C),
\end{equation}
where $\vr_C$ is the latent representations that encode information from input pairs as prior knowledge.
$p(\vy_i|\vx_i,\vr_C)$ is a likelihood that can be modelled by a Gaussian distribution factorised on target pairs $(\vx_i,\vy_i)_{i \in T}$ with the same distribution learning on $\vx_i$ and $\vr_C$. 
Beside the above deterministic path, another latent path with an encoder that extracts a latent representation $\vz$ from a learned distribution $\vs_C \coloneqq s(x_j, y_j)_{j\in C}$ to measure the uncertainty of output $\vy_i, i\in T$. Then, we have
\begin{equation}
    p(\vy_i|\vx_i,\vx_C,\vy_C) \coloneqq \int  p(\vy_i|\vx_i,\vr_C,\vz) q(\vz|\vs_C) d\vz,
\end{equation}
where $\vz$ represents different realizations of the stochastic process. 
The parameters in the deterministic path and latent path are trained on optimizing the following variational lower bound, which is also called evidence lower bound (ELBO):
\begin{equation}
    \label{eq:loss}
    \begin{aligned}
        \mathcal{L}_D&(\vx_T,\vx_C,\vy_C) = \log p(\vy_i|\vx_i,\vx_C,\vy_C) \geq\\
        &\E_{q}  [\log p(\vy_i|\vx_i,\vr_C,\vz)] -
        \KL ( q(\vz|\vs_T) \Vert q(\vz|\vs_C) ).
    \end{aligned}
\end{equation}
$D_{KL}$ measures the distance between the distribution learned on the target set and the context set. Practically, $C$ is a subset of $T$ to maintain the consistency of the reconstruction.
The desired model is required to meet the following principles: (i) Flexibility: define a rich family of distributions, where an arbitrary number of targets can be predicted conditioning on an arbitrary number of context samples. (ii) Scalability: computation cost scales linearly for points sampled during training and inference. (iii) Consistency: the predictions of the targets are invariant in the order of context samples.
However, it comes out that the existing methods are trapped with the problem that taken randomly sampled context points as individual elements, the dependency between the samples is disassembled and intractable.
Simply relying on the previous linear mapping mechanism is still far from fitting the target data distribution.

\subsection{Variational Auto-encoder}

The primary purpose is to learn generative function $p_{\theta}(\vx, \vz) = p_{\theta}(\vx\vert\vz)p_{\theta}(\vz)$ and an inference function $q_\phi(\vz|\vx)$ that approximates its posterior $p_\theta(\vz|\vx)$. The problem is that there is generally no guarantees about what these learned distributions actually are. 
We target at learning models satisfying the following implication which holds for all $(\vx, \vz)$:
\begin{align}
    \forall (\theta,\theta'):\;\; p_{\theta}(\vx) = p_{\theta'}(\vx) \;\;\implies\;\; \theta = \theta'.
    \label{eq:identifiable}
\end{align}
In other words, if any two different groups of parameter $\theta$ and $\theta'$ lead to the equal marginal density $p_\theta(\vx)$, this could imply that they have matching joint distributions $p_\theta(\vx,\vz)$. This means that if we learn a model with parameter $\theta$ that fits the data perfectly: $p_{\theta}(\vx) = p_{\theta^*}(\vx)$ , then its joint density also matches properly: $p_{\theta}(\vx, \vz) = p_{\theta^*}(\vx, \vz)$, which also means that we found the prior $p(\vz) = p_{\theta^*}(\vz)$ and posteriors $p(\vz\vert\vx) = p_{\theta^*}(\vz\vert\vx)$ correctly. On the other side, the inference model $q(\vz|\vx)$ can also be used to efficiently perform inference over the sources from which the data originates. 

However, a general problem for current methods is that they only consider mapping the global distribution while neglecting the importance of data relativity which ensures the model resistant to noise factors and observation bias. 
In this case, we can always find transformations of $\vz$ that change the values without interfering with the distribution. Take a spherical Gaussian distribution $p(\vz)$ for example, applying rotation keeps its distribution the same. We can then incorporate this transformation in $p(\vx\vert\vz)$. This will not change $p(\vx)$, but it will affect $p(\vz\vert\vx)$, since now the values of $\vx$ come from different values of $\vz$. 
The backbone structure of the proposed method is illustrated in Figure~\ref{fig:framework1}, which simultaneously learns a deep latent generative model and a variational approximation $q(\vz\vert\vx,\vu)$ of its posterior $p_{\theta}(\vz\vert\vx,\vu)$, the latter being often intractable. 

The conditional marginal distribution of the observations is denoted by $p(\vx\vert\vu) = \int p(\vx, \vz, \vert\vu)d\vz$, and with $q_{\mathcal{D}}(\vx, \vu)$ we denote the empirical data distribution given by dataset $\mathcal{D}$. The goal of VAEs is learn the vector of parameters $(\theta, \phi)$ by maximizing $\mathcal{L}(\theta, \phi)$, a lower bound on the data log-likelihood defined by:
\begin{equation}
\label{eq:loss}
    \begin{aligned}
        \E_{q_{\mathcal{D}}}&\left[ \log p(\vx\vert \vu)\right] \geq \mathcal{L} (\theta, \phi) := \\
        &\E_{q_{\mathcal{D}}} \left[ \E_{q(\vz \vert \vx,\vu)} \left[\log p(\vx,\vz\vert\vu) - \log q(\vz\vert\vx,\vu)\right]\right].
    \end{aligned}
\end{equation}
Then we apply the reparameterization trick \citep{kingma2015variational} to sample from $q(\vz\vert\vx,\vu)$. This trick provides a low-variance stochastic estimator for gradients of the lower bound with respect to $\phi$. Estimates of the latent variables can be obtained by sampling from the variational posterior. The training algorithm is the same as in a regular VAE. 


\subsection{Pointwise Relational Inference}
To exploit the data relativity within sample points, we propose to build connections inner the context points as well as cross the context and target points for relational inference.
Specifically, our method targets at learning representations that capture both the appearance features and data relationship which relies on position-aware embedding.
Instead of learning the point-to-point mapping with linear layers, we propose to formulate the data points into a graph structure to better learn the relativity among the data points. 
The pipeline and the details of OR-Net are illustrated in Figure~\ref{fig:framework2}. 
OR-Net is elaborated to learn the representation invariant to the order of samples while maintains the structural features which conform with the principles of exchangeability and consistency.

\noindent\textbf{Omni-Relational Layer.}
The Omni-Relational Layer is designed to learn the data relativity among the nodes in the input graph, and can be transformed into Inner-Relational Layer and Cross-Relational Layer accordingly. 
The embedding of the relative geometric information enables each node to better capture its relationship to its neighbors. A node embedding is position-aware if it fits in a function conditioned on its n-hop neighborhood.
At first, given a point set $S$, each sample $(\vx_i, \vy_i)_{i\in S}$ represents one node in the graph. $S$ is a subset of the context/target set. To embed the location information, we set a fixed value of radius as a distance threshold $\gamma$ to define the neighborhood for each node. 
The relative position-aware embedding is learned with MLP layers. The representation of each node is a combination of the node embedding with the geometric embedding. 

The key insight of OR-Net is that the node location can be captured by quantifying the relative distances between a point and a set of context points. 
To compute the geometric embedding, we first calculate the relative location between each sample with its neighbors, then learn from the distances $\{d^m\}_{m=0}^M$ along the $M$ dimensions of the position between node $v_i$ and $v_j$. $\bar{d}$ denotes the distance between two points. $M=2$ when points are sampled from a 2D coordinate.
Note that the values of location are normalized before calculation. The relative geometric embedding $\mathbf{r}_{ij}$ can be learned for each neighbor $j$ of node $i$ as following,
\begin{equation}\label{eq:pos_emb}
    \mathbf{r}_{ij} = W_d \textsc{concat}[\{d^m_{ij}\}_{m=0}^M, \bar{d}],
\end{equation}
where $W_d$ is a learnable weight matrix that transforms the distances into a vector. Combining node features and geometric embedding can then be concatenated.
Geometric information is critical to reveal the relationship between each node with its neighbors, while the node features provide side information that is useful for the prediction task.

\noindent\textbf{Inner-Relational Learning in the Context.} 
Given the context points from the partial observation, we first build a context graph $G_c$ to learn the inner relationship. We assume the message computation function $\mathcal{F}(i,j,\mathbf{h}_j, \mathbf{r}_{ij})$ accounts for both the node features as well as geometric embeddings. The following function $\mathcal{F}$ performs well for learning the node embedding parameterized with $W_h$:
\begin{equation}\label{eq:message}
   \mathcal{F}(i,j,\mathbf{h}_j, \mathbf{r}_{ij}) = W_h\textsc{Concat}(\mathbf{h}_j,\mathbf{r}_{ij}).
\end{equation}

The node feature is learned by concatenating the features of its neighbors, similar to \cite{hamilton2017inductive}, or by passing in the information from the neighboring nodes. The former can be realized by learning attention computed by a learnable linear transformation. Inspired by~\cite{velivckovic2017graph}, we use the attention mechanism to retrieve useful messages from the neighbors for each node. As an initial step, a shared linear transformation is applied to every node. Then, self-attention is performed on the nodes, a shared attentional mechanism computes attention coefficients $a_{ij}$ 
which indicates the importance of the embedding of node $j$ to node $i$. In its most general formulation, the model allows every node to attend to every other node, dropping all structural information. 

In this way, we inject the graph structure into the mechanism by calculating attention as $a_{ij}$ for nodes $j\in\mathcal{N}_i$, where $\mathcal{N}_i$ is the neighborhood of node $i$. 
The use of dot-product attention allows the query values to be computed with two matrix multiplications and a softmax, allowing for use of highly optimised matrix multiplication code. 
Thereafter, the normalized attention coefficients are used to compute a linear combination of the features corresponding to them, to serve as the final output features for every node:
\begin{equation}\label{eqnatt}
	\hat{h}_i = \sum_{j\in\mathcal{N}_i} a_{ij} W_a h_j,
\end{equation}
where $h_j$ is learned embedding from the message computation function $\mathcal{F}$ in Eq.~\ref{eq:message}.
Once the attention weights are learned, we normalize them across all neighbor nodes using the softmax function to make coefficients easily comparable across different nodes.

\noindent\textbf{Cross-Relational Inference for the Target.} 
We further investigate building the relationship cross the context points and the target data points. It is noted that we include the context points into the target points in this step. So, a full graph $G_t$ is built upon both the context points and target points.
Similarly, we first embed both the node features and geometric features with $\mathcal{F}$ to compute the messages.
Then, the attention coefficients $a_{ij}$ are calculated and normalized accordingly. Each node embedding is updated by aggregating the messages from its neighbors.
By passing messages across Omni-Relational Layers, each node retrieves messages from its multi-hop neighbors in a large receptive field. After the message aggregation, the updated node features are adopted to learn the deterministic feature $\vr$ and the latent feature $\vz$.

\subsection{Objectives}

The previous methods only focus on the maximization of the mutual information $I$ between the context points with the data distribution. However, the learned distribution is variant to the bias brought by the limited observation and thus causes unstable and unreliable estimation. 
It is widely adopted as a critical principle for representation learning that an optimal representation should be composed of minimal sufficient information. It encourages the representation to be informative and compact about the target to make the prediction satisfied. On the other hand, the information bottleneck (IB) also discourages the representation from acquiring additional information from the data that is irrelevant for predicting the target (minimal). Motivated by this paradigm, the learned model tends to avoid  over-fitting and becomes more robust and reliable to the uncertainty of estimation by minimizing the function:
\begin{equation}
    \mathcal{L}_{\mathrm{IB}}(\vx,\vz; \beta)=-I(\vz; \vy)+\beta I(\vx; \vz),
\end{equation}
where $\beta$ is a balancing parameter for the loss terms.
Inherited from the IB principle, each node embedding is optimized by minimizing the information from the conditional context and maximizing the information from its original distribution.
In order to better incorporate IB with the graph-based structure, here we follow the local dependence assumption to constrain the optimizing space: Messages aggregated on each node from its neighbors within n-hops. Hence, the objective of OR-Net could be reformulated as:
\begin{equation}
    \mathcal{L}= \mathcal{L}_D(\vx_T,\vx_C,\vy_C) + \mathcal{L}_{IB}(\vx_T,\vz; \beta).
\end{equation}

The proposed OR-Net transforms the randomly sampled point set into structural data with the constrain of the neighborhood of each point. Beyond that, the embedded geometric information assists in building the relationship between each target point with the context point set through the neighbors. With the geometric attentive message passing and aggregation, both of the appearance embedding and geometric embedding are passed into the next layer. It is further proved empirically that representation learned by OR-Net is more expressive and reliable.

\section{Experiment}

\subsection{Training Details}
\noindent\textbf{Network structure.} OR-Net is built upon encoder-decoder structure. Specifically, the encoder is composed of a deterministic path to learn the representation and a latent path to estimate the data distribution. It is worth noted that the same weights are shared in the latent path for learning both the prior distribution on the observed data points and the posterior distribution. In the Omni-Relational Layer, attentive pooling functions as the message passing function. Then, a three-layer MLP functions as the decoder to predict the concrete values for the target points. Optimization is performed with Adam~\cite{kingma2014method}. 
We set batch size for training to be 32 for function regression and image completion, 128 for motion generation. The learning rate is initialized to be 0.001.
The balancing parameter $\beta$ is set to 0.05 empirically in our experiment.

\noindent\textbf{Comparison methods.} For a comprehensive comparison, we conduct experiments for conditional generation of different modalities against Gaussian Processes~\cite{williams1996gaussian}, Conditional NP~\cite{garnelo2018conditional}, Attentive NP~\citep{kim2019attentive}. Additionally, we also implement a graph-based method PGNN~\cite{you2019position}. 
During training, our model is built upon graphs with a fixed node ordering and re-trained invariant to the ordering of the nodes. We report the test set performance when the best performance on the validation set is achieved, and results are reported over 5 runs with random seeds.

\subsection{1D Function Regression}
We start with conducting experiments for 1D function regression task.
As a stochastic process, OR-Net is trained by sampling a batch of realizations from the data generating where random points are selected to be the targets and a subset to be the context set to approach the target points. We first explore data generated from a Gaussian Process with a squared-exponential kernel and small likelihood noise. The number of contexts (n) and number of targets (m) are chosen randomly at each iteration, (n $\sim$ U[3, 20], m $\sim$ n + U[0, 20 $-$ n]). Each data point is drawn uniformly with random choice for $\vx$ in (-2, 2). Each sequence contains 100 points.

\begin{figure}[t]
  \centering
  \includegraphics[width=\columnwidth]{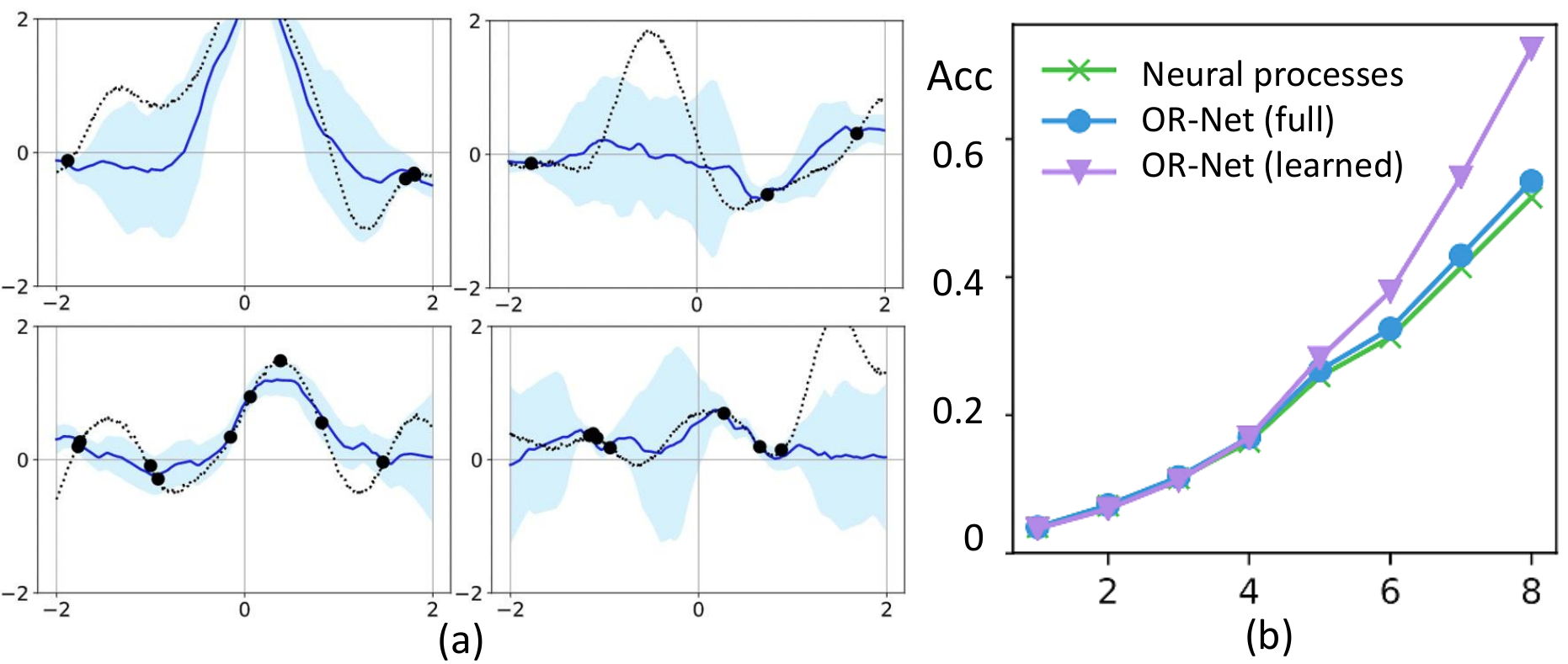}
  \caption{
    We present the training and inference results for 1D function regression. Specifically, we show the predicted results with blue lines in (a) for function regression where the context points are drawn with dots and the ground truth are grey dotted lines, and also the evaluated accuracy with varying context numbers in (b). 
  }
  \label{fig:demo_1d}
\end{figure}

We show the performance for function regression in Figure~\ref{fig:demo_1d}. The estimated regression results are shown in blue lines conditioned on the context points drawn with dots. It is observed on the right that OR-Net performs better compared to Neural Processes. Furthermore, with the learned data structure, OR-Net can better adapt to the observed data points with higher accuracy. When supplied with less context points, NPs have incorrectly collapsed the distribution over functions to a set of almost horizontal lines. OR-Net, on the other hand, is able to produce a wide range of possible trajectories. Even when a large number of points have been supplied, the NP posterior does not converge on a good fit, while OR-Net correctly captures the latent variables. 

\begin{table}
  \centering 
  \caption{We evaluate methods on MNIST with context points number in (50, 100, 200, 400). OR-Net$\dagger$ indicates OR-Net is trained with IB loss.}
  \setlength{\tabcolsep}{1.25mm}
  \begin{tabular}{lcccc}
    \toprule
         & 50    & 100  & 200  & 400  \\
    \midrule
    CNP~\cite{garnelo2018conditional}       & $0.073\small{\pm .013}$ & $0.036\small{\pm .006}$ & $0.022\small{\pm .005}$ & $0.024\small{\pm .006}$ \\
    PGNN~\cite{you2019position}     & $0.060\small{\pm .010}$ & $0.032\small{\pm .006}$ & $0.025\small{\pm .007}$ & $0.022\small{\pm .003}$ \\
    ANP~\cite{kim2019attentive}       & $0.056\small{\pm .010}$ & $0.029\small{\pm .004}$ & $0.015\small{\pm .003}$ & \textbf{0.009$\small{\pm .003}$} \\
    \midrule
    OR-Net       & $0.049\small{\pm .005}$ & $0.025\small{\pm .004}$ & $0.012\small{\pm .004}$ & $0.010\small{\pm .003}$ \\
    OR-Net$\dagger$       & \textbf{0.045$\small{\pm .006}$} & \textbf{0.022$\small{\pm .003}$} & \textbf{0.010$\small{\pm .004}$} & \textbf{0.009$\small{\pm .002}$} \\
    \bottomrule
  \end{tabular}
  \label{tab:1-mnist}
\end{table}

\subsection{Image Completion}

Images are composed of arranged points that can be interpreted as being generated from a stochastic process.
The performance of the proposed OR-Net is compared quantitatively with the former approaches, kNN, GP~\cite{williams1996gaussian}, NP~\cite{garnelo2018conditional}, ANP~\cite{kim2019attentive} and also our realization of PGNN~\cite{you2019position}. 
Each image is considered as one realization sampled on a fixed 2-dimensional grid. 
All experiments are trained up to 200 context/target points at training with a random number at each iteration. 
Neural processes aim to predict the value of target points which can be dealt with as a regression problem.
Given the context points as pairs of $\{\vx_j, \vy_j\}_{j\in C}$, the objective is to map the location of a target point $\vx_i$ to its pixel value $\vy_i\in \R^1$ as greyscale, $\vy_i\in \R^3$ as RGB. 

We conduct experiments on MNIST~\cite{lecun1998gradient} and CelebA~\cite{liu2015deep} with the default train/test splits. We resize the MNIST images as $28 \times 28$. For CelebA, we crop the images as $128 \times 128$ and then resize images as $32 \times 32$. Empirically, the number of graph layers is set as 2 in our experiments. 
We first compare models during inference with different numbers of context points on MNIST dataset. Models are uniformly trained with less than 200 context points. It is also observed that OR-Net performs better with fewer context points during the inference with results in Table~\ref{tab:1-mnist}. Both insights show the fact that with the geometric graph, OR-Net takes the advantage of the position-aware representation to better fit the data distribution.
The results on the CelebA dataset are summarized in Table~\ref{tab:2-celeba} with an increasing number of context points in (10, 100, 1000) sampled either at random or ordered from the top-left corner.

\begin{table}
  \centering
  \caption{Mean square error measured pixel-wise for image completion task on CelebA. Performance of experiments with random and ordered (top-left) context points of (10, 100, 1000) are compared. }
  \setlength{\tabcolsep}{2mm}{
  \begin{tabular}{lllllll}
    \toprule
    & \multicolumn{3}{c}{Random} & \multicolumn{3}{c}{Ordered} \\
    \cmidrule(r){2-7}
    Model     & 10  & 100  & 1000  & 10  & 100  & 1000  \\
    \midrule
    kNN       & $0.215$ & $0.052$ & $0.007$ & $0.370$ & $0.273$ & $0.007$   \\
    GP~\cite{williams1996gaussian}       & $0.247$ & $0.137$ & \textbf{0.001} & $0.257$ & $0.220$ & \textbf{0.002}      \\
    PGNN~\cite{you2019position}     & $0.057$ & $0.026$ & $0.015$ & $0.122$ & $0.049$ & $0.017$ \\
    CNP~\cite{garnelo2018conditional}       & $0.039$ & $0.016$ & $0.009$ & $0.057$ & $0.047$ & $0.021$ \\
    ANP~\cite{kim2019attentive}       & $0.032$ & $0.013$ & $0.006$ & $0.044$ & $0.031$ & $0.012$ \\
    \midrule
    OR-Net       & $0.024$ & $0.011$ & $0.008$ 
              & $0.039$ & $0.027$ & $0.014$ \\
    OR-Net$\dagger$    & \textbf{0.021} & \textbf{0.008} & 0.006
              & \textbf{0.036} & \textbf{0.024} & 0.011 \\
    \bottomrule
  \end{tabular}}
  \label{tab:2-celeba}
\end{table}

\begin{figure}[t]
  \centering
  \includegraphics[width=\columnwidth]{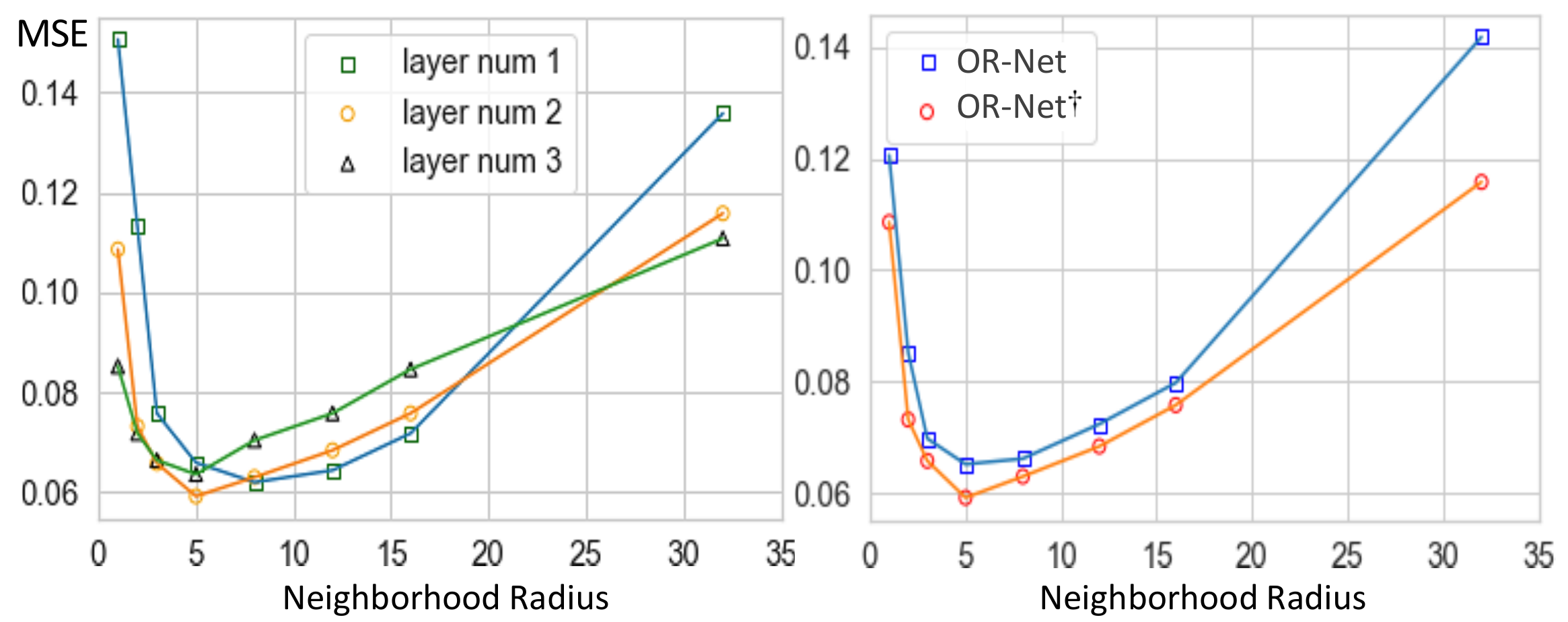}
  \caption{
    We report our results with different numbers of Omni-Relational Layer in OR-Net and the hyper-parameter neighborhood radius $\gamma$ in the graph structure in the left. The right part compares the performance of OR-Net with and without IB loss.
  }
  \label{fig:ablation}
\end{figure}

We further executed grid search with different values of neighborhood radius $\gamma$ in range of [1, 2, 3, 5, 8, 12, 16, 32] and graph layer number in [1, 2, 3] to further explore the position embedding with results shown in Figure~\ref{fig:ablation}. 
It is discovered that the performance reaches the best when $\gamma$ is set to 5 with 2 geometric graph layers. The performance drops when the value of $\gamma$ is too small or too large. We also applied different values of neighborhood radius as the percentage against the image size to further explore the position embedding. The performance reaches the best at 0.5. Thus, with a modest receptive field, OR-Net can better capture the data structure in a hierarchical way.
It is also shown that OR-Net performs consistently better with the constrain of information bottleneck.

\begin{figure}[t]
  \centering
  \includegraphics[width=0.9\columnwidth]{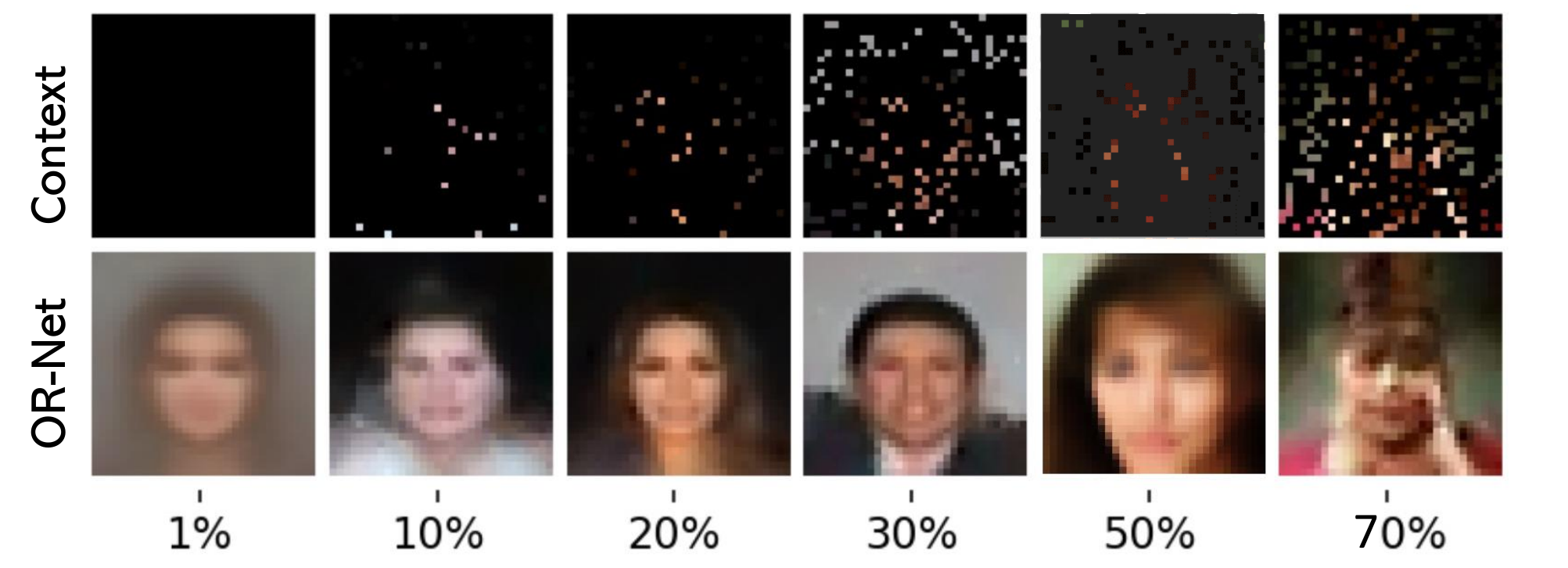}
  \caption{
    Reconstruction results by OR-Net on the test set of CelebA with context points with different percentages.
  }
  \label{fig:2d_celeba}
\end{figure}

\begin{figure}[t]
  \centering
  \includegraphics[width=1.05\columnwidth]{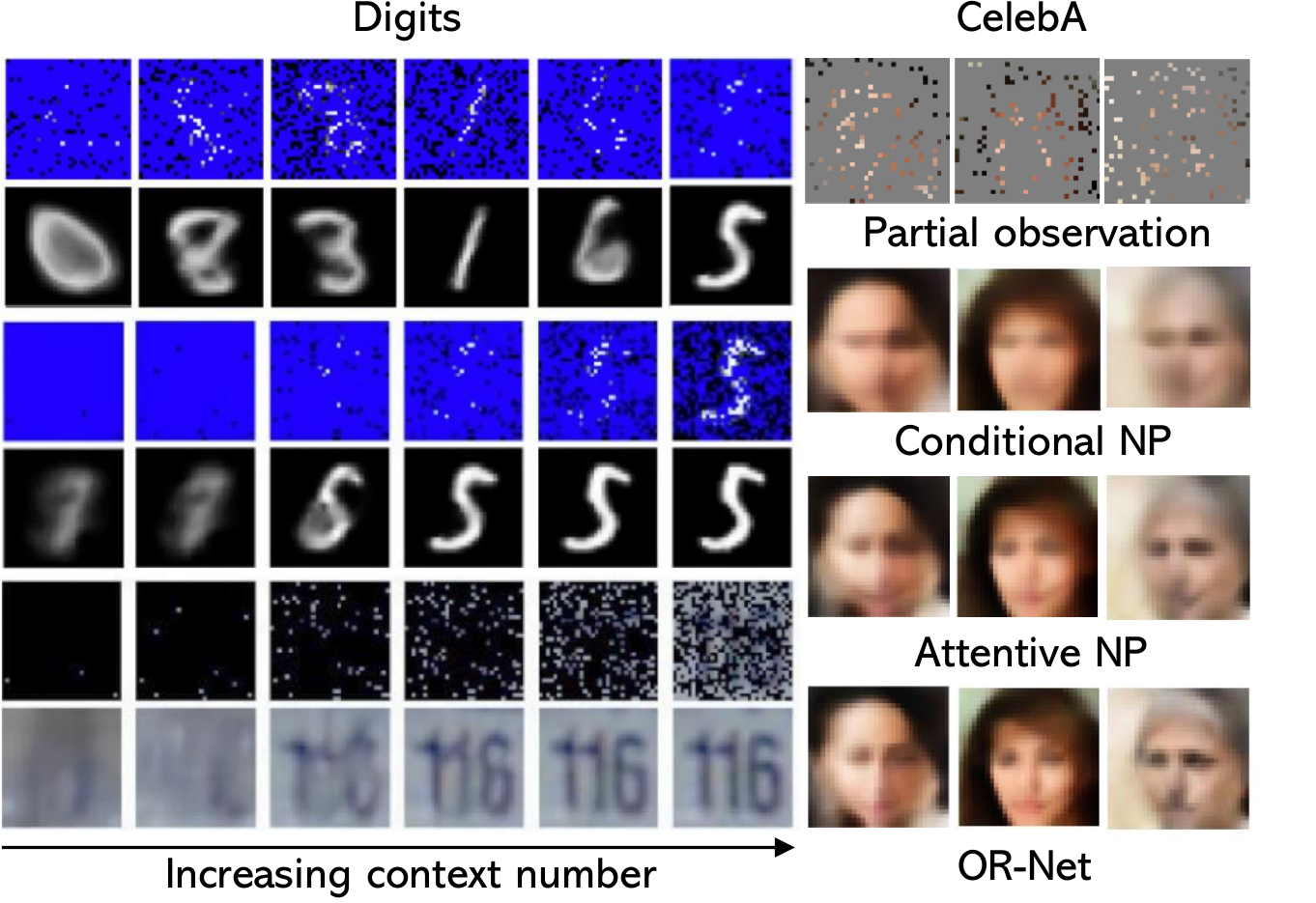}
  \caption{
    Reconstruction results on digit dataset: MNIST and SVHN with increasing context points on the left, and reconstruction on CelebA by Conditional NP, Attentive NP and OR-Net (with only 10\% points are sampled as context).
  }
  \label{fig:2d_digit}
\end{figure}

In Figure \ref{fig:2d_celeba}, we show predictions of the full image pixels as target points at different training stages chronologically for randomly selected images. 100 context points are sampled for the inference. It is observed that OR-Net gives reasonable predictions with better reconstructions in the fine-grained details, \eg, the edge of face, the shape of nose and eyes. The embedding of position information also helps achieve more smooth and natural image completion results compared to the other methods. It is also proved that OR-Net can better model the global structure of the image, with one sample corresponding to one realization of the data generating stochastic process. We also visualized the reconstruction of digits with increasing context number in the left of Figure~\ref{fig:2d_digit} and compared the face reconstruction for images in CelebA against other methods.
When sampling from this model with a small number of observed pixels, we get coherent samples and we see that the variability of the datasets is captured. As the model is conditioned on more and more observations, the variability of the samples drops and they eventually converge to a single possibility. It is demonstrated that by exploiting the potential pointwise relationship in modalities without observed structure, OR-Net can better model the latent variables and underlying distribution.

\begin{figure*}[t]
  \centering
  \includegraphics[width=2\columnwidth]{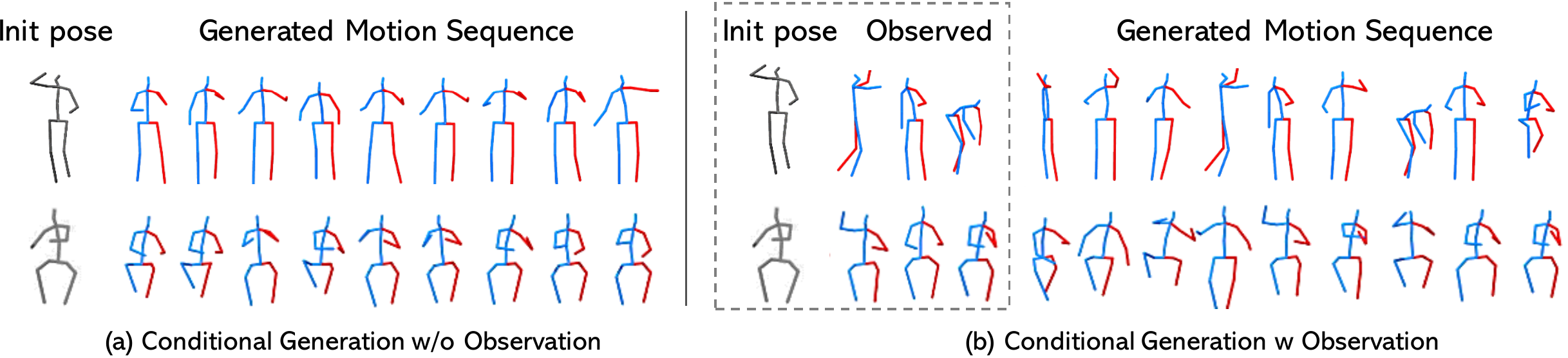}
  \caption{
    Reconstruction sequences for action ``walking'' and ``sitting'' in HumanAct12 without observed poses on the left. On the right, we show the reconstructed results given 3 poses randomly sampled as the context set.
  }
  \label{fig:3d_demo}
\end{figure*}

\subsection{Sequential Motion Generation} 
Motion capture equipment is widely adopted to generate realistic and smooth motion. However, the motion capture process is expensive and time-consuming. These drawbacks require the original data to be well labeled and further processed. To address these issues, we extend conditional generation into the motion completion based on HumanAct12~\cite{guo2020action2motion}, an in-house dataset which is adopted from an existing dataset PHSPD~\cite{zou2020polarization}, consisting of 1,191 motion clips and 90,099 frames in total, with hierarchical action type annotations. 
HumanAct12 has more organized action annotation, with a balanced number of motions per action compared to other datasets. A body pose consists of 24 joints (23 bones). 

We show in Figure~\ref{fig:3d_demo} that given 3 randomly sampled poses from a sequence, our method is able to generate natural and meaningful motion sequences. We use conditional GRU as our deterministic baseline. Vanilla RNN model takes the condition vector and pose vector together as input at current step and predicts the pose vector for the next step. For conditional GRU, generated poses all collapse to a set of spatial points near the root joint, which shows the inefficacy of simple RNN models toward a non-deterministic generative task. It is worth noting that motion generation should not be a one-to-one mapping process. Instead, the generated motions are expected to be close to the real motions in terms of their respective distributions.

The most notable difference from the aforementioned tasks is that human motion is generated upon visible skeleton structure. Additionally, we use LSTM in preserving the motion variance in the time dimension. We investigate how well OR-Net can capture the latent patterns in different motions, using several types of motions with different properties. We first train OR-Net on them separately then jointly. We observe that OR-Net can learn the transition stochasticity well when trained on a single type of motions. The diversity can be found in short-term and long-term transitions, which are two levels of multi-modality captured well by OR-Net. The action-level transition has also been captured and generated by conditioning on the given action label. The reconstruction sequences for action ``walking'' and ``sitting'' are presented in Figure~\ref{fig:3d_demo}. We can observe that with the given context, the generated motion sequences can be more diverse with the required poses. Similar observations are also found in other motions.

\begin{table}
  \centering 
  \setlength{\tabcolsep}{5mm}
  \caption{The ablation of different modules in Omni-Relational Network reported with MSE on MNIST given 10$\%$ context points by ablating (1) graph structure (2) attentive pooling (3) positional embedding (P.E.) (4) information bottleneck constrain (I.B.).}
  \begin{tabular}{lcccc}
    \toprule
      Graph       & Attention       & P.E.            & I.B.             & MSE \\
    \midrule
                  &                 &                &                   & $0.073$\\
    $\checkmark$   &              &                 &                    & $0.066$\\
    $\checkmark$  &               &               &    $\checkmark$    & $0.058$\\
    $\checkmark$  &                 &   $\checkmark$     &  $\checkmark$   & $0.053$\\
    $\checkmark$  &  $\checkmark$  &               &     $\checkmark$   & $0.049$\\
    $\checkmark$  &  $\checkmark$  &   $\checkmark$    &  $\checkmark$   & $0.045$\\
    \bottomrule
  \end{tabular}
  \label{tab:3-ablation}
\end{table}

\begin{figure}[t]
  \centering
  \includegraphics[width=\columnwidth]{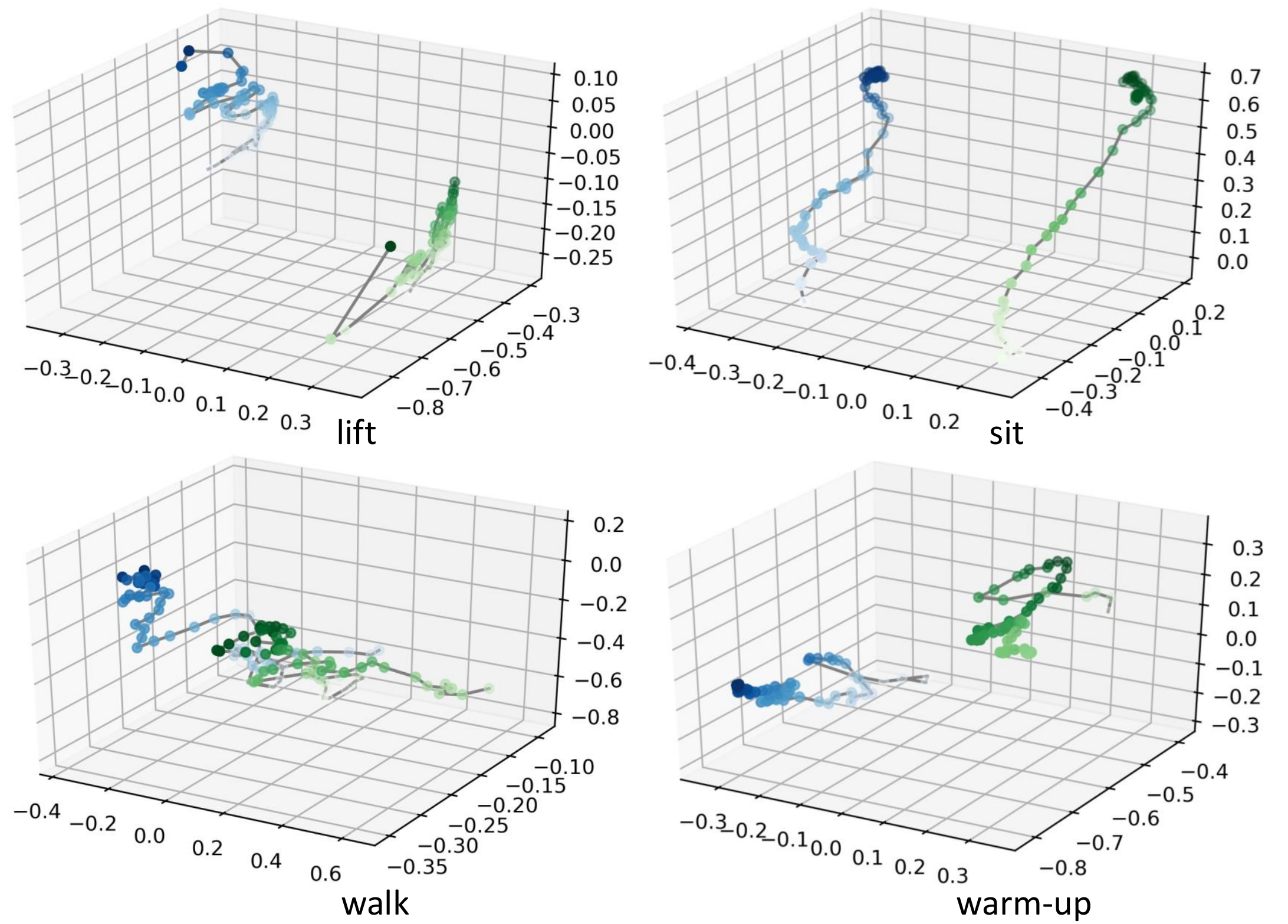}
  \caption{
    The visualized trajectories for joint 18 in the left arm and joint 19 in the right arm for different action classes. It can be observed that the correlation is high in ``sit'' and ``walk'' but not for the other two actions.
  }
  \label{fig:ablation_nb}
\end{figure}


Furthermore, in table~\ref{tab:3-ablation}, performances are reported with the ablation of the sub-modules in OR-Net: (1) graph structure (2) attentive pooling (3) positional embedding (P.E.) (4) information bottleneck constrain (I.B.). It is observed that with the relational inference in the graph structure, OR-Net reaches considerable results. Additionally, the information bottleneck constrain further boosts the performance steadily.

\section{Conclusion}

In this paper, we explore a new scheme, Omni-Relational Network, to derive a more robust and general representation of the data. In this regard, OR-Net exploits the data relativity within not only the context samples but also cross the context and the target points by fusing the positional embedding. Simultaneously, it retains the benefits of information bottleneck to enhance the compactness of representation.
By transforming the randomly sampled data into a position-based structure, we offer a promising direction to dig into the potential relevance under the partial observation.
Furthermore, this work can be seen as a step towards learning high-level abstractions of the data distribution with fewer samples or incomplete data.
In future work, we will explore how far OR-Net can help to tackle learning problems hinge on the inherited relationship in data of different modalities, such as meta-learning and data efficiency.

{
\bibliographystyle{ACM-Reference-Format}
\bibliography{acmart}
}

\end{document}